\documentclass[11pt]{article}

\usepackage[left=2.1cm, right=2.1cm, top=2cm, bottom=2cm]{geometry}

\usepackage[utf8]{inputenc}
\usepackage[T1]{fontenc}
\usepackage{amsfonts,amsmath,amssymb,amsthm}
\usepackage{microtype}
\usepackage{graphicx}
\usepackage{subcaption}

\usepackage{natbib}
\usepackage{hyperref}
\usepackage{url}
\usepackage[capitalize,noabbrev]{cleveref}

\usepackage[dvipsnames,svgnames]{xcolor}

\hypersetup{
    colorlinks=true,
    linkcolor=blue!50!black,
    citecolor=blue!50!black,
    filecolor=magenta,
    urlcolor=cyan
}

\usepackage{authblk}

\newtheorem{proposition}{Proposition}
\newtheorem{theorem}{Theorem}
\newtheorem{corollary}{Corollary}

\theoremstyle{remark}

\newcommand{\E}{\mathbb{E}}
\newcommand{\N}{\mathcal{N}}
\newcommand{\Id}{\mathrm{I}}
\newcommand{\Law}{\mathrm{Law}}

\title{Gaussian Flow-Matching Schedules:\\
Implications for Sampling and Training}

\author[1]{Arsène Claustre\textsuperscript{*}}
\author[1,2]{Hugo Negrel\textsuperscript{*}}
\author[3,4]{Claire Boyer}
\author[1]{Kimia Nadjahi}
\author[1,2,5]{Eric Vanden-Eijnden}

\affil[1]{CNRS, ENS Paris}
\affil[2]{Machine Learning Lab, Capital Fund Management, Paris, France}
\affil[3]{Laboratoire de Mathématiques d'Orsay, CNRS, Université Paris-Saclay}
\affil[4]{Institut Universitaire de France}
\affil[5]{Courant Institute of Mathematical Sciences, New York University}

\affil[ ]{\textsuperscript{*}Equal contribution}

\date{}

\begin{document}
\maketitle

\begin{abstract}
Flow-matching schedules affect both sampling dynamics and the variance of the regression target. For centered commuting Gaussians, we show that a direction-dependent schedule decomposes into two independent design choices: a variance path, which fully determines the intermediate laws and probability flow, and a factorization, which leaves this flow unchanged while controlling irreducible regression variance. On the sampling side, we analyze finite-step Euler accuracy and derive a necessary drift bound for exact $N$-step sampling, connecting the geodesic and the logarithmic path. On the training side, for any fixed path, we derive closed-form factorizations that either minimize time-averaged regression variance or make it constant along the path.
\end{abstract}

\section{Introduction}

Flow matching constructs a transport from a simple source distribution $\mu_0$ to a target distribution $\mu_1$ \citep{albergo2022,albergo2023,lipman2022,liu2022}. Given endpoint samples $X_0\sim\mu_0$ and $X_1\sim\mu_1$, for $t\in[0,1]$ define the stochastic interpolant 
\begin{equation}
\label{eq:interp}
    I_t=\alpha_tX_0+\beta_tX_1,
\qquad
(\alpha_0,\beta_0)=(\Id_D,0),\quad
(\alpha_1,\beta_1)=(0,\Id_D),
\end{equation}
where $\Id_D$ is the identity matrix of size $D$, and $(\alpha_t,\beta_t)$ is a schedule determining how $X_0$ and $X_1$ are mixed along the path. 

Flow matching learns the velocity field $b_t(x)=\E[\dot I_t\mid I_t=x]$
by regressing the interpolant velocity $\dot I_t$ against its current state $I_t$. 
At inference, samples are generated by integrating the learned ODE $\dot X_t=b_t(X_t)$ with $X_0\sim\mu_0$.
While the endpoint constraints in~\eqref{eq:interp} ensure that the interpolation connects $\mu_0$ to $\mu_1$, they leave the intermediate schedule $(\alpha_t,\beta_t)$ unspecified. Choosing this schedule is therefore a key design choice, affecting both inference and training. At inference, it determines the probability path, hence the numerical complexity of discretizing the ODE. At training, it determines the regression target $\dot I_t$ and its conditional regression variance $\operatorname{Var}(\dot I_t\mid I_t)$. These two objectives need not be aligned: a schedule yielding a simple probability path may still induce a noisy regression problem.

Existing schedule-design strategies primarily target sampling efficiency, through kinetic or Jacobian criteria \citep{shaul2023kineticoptimalprobabilitypaths, cvx2025}, Lipschitz regularity \citep{tsimpos2025}, or few-step discretization accuracy \citep{karras2022,aranguri2025}. More recently, \citet{hurault2026geometryawarediscretizationerrordiffusion} optimize diffusion schedules by directly minimizing the leading-order SDE discretization error, with explicit solutions for Gaussian targets.

These approaches optimize the path of intermediate distributions and its sampling dynamics. We show that this does not exhaust the design of a stochastic interpolant. In each direction, the schedule decomposes into a \emph{variance path} $r_t$, which determines the intermediate distributions and probability flow ODE, and a \emph{mixing angle} $\theta_t$, which leaves this flow unchanged, but controls the conditional regression variance $\operatorname{Var}(\dot I_t\mid I_t)$. This separates schedule design into two complementary components: the path controls sampling, while its factorization controls the intrinsic difficulty of the flow-matching regression.

Our contributions are threefold: we establish the path-factorization decomposition, exhibit necessary conditions for exact finite-step Euler sampling, and derive closed-form factorizations that minimize or equalize the conditional regression variance along any fixed path. 

\section{Path-factorization decomposition}\label{sec:decomp}

We first consider the Gaussian case, where the interpolation schedule separates into two components controlling sampling and training, respectively.
Let $\mu_0=\N(0,C_0)$ and $\mu_1=\N(0,C_1)$, with $C_0, C_1\in\mathbb{R}^{D\times D}$ positive-definite and commuting. In a common orthonormal eigenbasis $u_1,\ldots,u_D$, write 
\[
C_0u_i=c_i u_i,
\qquad C_1u_i=c_i\rho_i u_i,
\qquad c_i,\rho_i>0,
\]
where $c_i$ is the source variance in direction $u_i$, and $\rho_i$ the target-to-source variance ratio. We consider stochastic interpolants~\eqref{eq:interp} with independent $X_0\sim\mu_0$ and $X_1\sim\mu_1$, and matrix-valued schedules $\alpha_t,\beta_t$ diagonal in the same basis. Their entries $\alpha_{t,i}$ and $\beta_{t,i}$ are twice continuously differentiable and nonnegative,  with endpoint conditions $\alpha_{0,i}=\beta_{1,i}=1$, $\alpha_{1,i}=\beta_{0,i}=0$.
For each direction, we reparametrize the schedule as
\begin{equation}\label{eq:polar}
\alpha_{t,i}=\sqrt{r_{t,i}}\cos\theta_{t,i},\quad
\beta_{t,i}=\sqrt{r_{t,i}/\rho_i}\sin\theta_{t,i}, \quad \text{where} \quad r_{t,i}=\alpha_{t,i}^2+\rho_i\beta_{t,i}^2,
\end{equation}
where $\theta_{t,i} \in [0, \pi/2]$, with $\theta_{0,i}=0$ and $\theta_{1,i}=\pi/2$, so that $r_{t,i}$ is the variance path, 
while $\theta_{t,i}$ describes how the randomness of the source and target are mixed along this path. 

\begin{proposition}[Path-factorization separation]\label{prop:separation}
The flow-matching drift $b_t$ between $\mu_0=\mathcal{N}(0,C_0)$ and $\mu_1 = \mathcal{N}(0,C_1)$ is linear and diagonal in the common eigenbasis of $C_0$ and $C_1$, with
\begin{equation}\label{eq:drift}
b_t(x)_i=a_{t,i}x_i,
\qquad
a_{t,i}=\frac{\dot r_{t,i}}{2r_{t,i}}.
\end{equation}
Moreover, 
\begin{equation}\label{eq:variance}
\operatorname{Var}(\dot I_{t,i}\mid I_{t,i})
=c_i r_{t,i}\dot\theta_{t,i}^{2}.
\end{equation}
Consequently, the marginal path and the drift depend only on $r_{t,i}$, whereas $\theta_{t,i}$ controls the conditional regression variance without affecting either.
\end{proposition}
\Cref{prop:separation} is proved in Appendix~\ref{app:separation} and has a simple geometric interpretation. Let
\[
z_{t,i}=(\alpha_{t,i},\sqrt{\rho_i}\beta_{t,i})
    =\sqrt{r_{t,i}}(\cos\theta_{t,i},\sin\theta_{t,i})
\]
so that $I_{t,i} = \sqrt{c_i} \langle z_{t,i}, G \rangle$, $G \sim \mathcal{N}(0, I_2)$. The radius $\sqrt{r_{t,i}}$ determines the marginal variance
$\operatorname{Var}(I_{t,i})=c_i r_{t,i}$, thus the Gaussian marginal law. Its angle $\theta_{t,i}$ does not change the law but controls the component of $\dot z_t$ orthogonal to $z_t$ and hence the conditional regression variance
$\operatorname{Var}(\dot I_{t,i}\,|I_{t,i})$.

\section{The path controls inference}\label{sec:path}

We now examine how the variance path $(r_t)$ shapes the inference dynamics. We first compare standard continuous-time criteria for path design, then study their implications for numerical integration.

\textbf{Continuous-time criteria.} Let $\varphi_{t,i}=\sqrt{r_{t,i}}$. We consider two standard criteria: the kinetic energy $\mathcal{K}$, as in dynamic optimal transport, and the integrated squared Jacobian $\mathcal{J}$, which provides a smooth surrogate for the uniform spatial Lipschitz constant \(\mathcal L_\infty(b)
=\sup_{t\in[0,1]}\operatorname{Lip}(b_t)
=\max_i\sup_t|a_{t,i}|.\)
\begin{align*}
\mathcal K(b)
&=\int_0^1\E[|b_t(I_t)|^2]\,\mathrm{d}t
 =\sum_{i=1}^D c_i\int_0^1\dot\varphi_{t,i}^2\,\mathrm{d}t,\\
\mathcal J(b)
&=\int_0^1\E[\|\nabla_xb_t(I_t)\|_F^2]\,\mathrm{d}t
 =\sum_{i=1}^D \int_0^1|a_{t,i}|^2\,\mathrm{d}t. 
\end{align*}

These objectives have distinct optimal paths. For clarity, fix a direction $u_i$ and let $r_t=r_{t,i}$, $\rho=\rho_i$, $\varphi_t=\sqrt{r_t}$, with $\varphi_0=1$, $\varphi_1=\sqrt\rho$.
Then, $\mathcal K$ is minimized by the Gaussian Wasserstein-2 geodesic,
\begin{equation}\label{eq:straight}
\varphi_t=1-t+t\sqrt\rho,
\qquad
a_t=\frac{\sqrt\rho-1}{1-t+t\sqrt\rho},
\end{equation}
whereas $\mathcal J$ is minimized by the log-linear variance path,
\begin{equation}\label{eq:logpath}
r_t=\rho^t,
\qquad
a_t=\frac12\log\rho.
\end{equation}
Note that \eqref{eq:logpath} has constant drift and also minimizes $\sup_t|a_t|$, since
$\sup_t|a_t|
\geq
\left|\int_0^1a_t\mathrm{d}t\right|
=\frac12|\log\rho|$~\citep{cvx2025}. 

\textbf{Euler discretization.} These continuous-time criteria do not, however, directly characterize finite-step sampling accuracy. We therefore turn to the Euler discretization of the probability-flow ODE.
Let $X_t$ solve $\dot X_t=b_t(X_t)$. Spatial Lipschitz regularity controls the propagation of discretization errors, while the local Euler truncation error is controlled by the material derivative 
\(
\partial_t b_t+\nabla_xb_t\,b_t.
\) 
In our Gaussian setting, since $X_t$ and $I_t$ share the same marginal law, one has
\[
\mathcal A(b)
:=\int_0^1
\E\!\left[
|\partial_tb_t(I_t)+\nabla_xb_t(I_t)b_t(I_t)|^2
\right]\mathrm{d}t
=\sum_{i=1}^D c_i\int_0^1\ddot\varphi_{t,i}^2\,\mathrm{d}t\,.
\]
If $\operatorname{Lip}(b_t)\le L$ uniformly and $\widetilde X_1$ is obtained with an Euler discretization of the probability-flow ODE with a step size $h$, a synchronous coupling gives (Appendix~\ref{app:eulerbound})
\begin{equation}\label{eq:eulererror}
W_2\bigl(\Law(X_1),\Law(\widetilde X_1)\bigr)
\le \frac{e^L h}{\sqrt3}\, \mathcal A(b)^{1/2}.
\end{equation}
Thus, Euler error depends on stability through $L$, and on consistency through $\partial_tb_t+\nabla_xb_tb_t$.

This bound separates the straight and logarithmic paths from the viewpoint of discretization. 
For the logarithmic path, the drift is constant in time but the trajectory is exponential,
\(
\varphi_t=\rho^{t/2}.
\)
Thus $\partial_t b_t=0$ does not imply zero local truncation error: the material derivative remains
\[
\partial_t b_t(X_t)+\nabla_x b_t(X_t)b_t(X_t)
=a_t^2X_t.
\]
Consequently, Euler does not integrate the logarithmic path exactly at finite step size.

By contrast, the straight path has a time-varying drift but an affine flow multiplier, so that
\(
\varphi_t=1-t+t\sqrt\rho\), and then \(\ddot\varphi_t=0.\)
Its material derivative therefore vanishes along the probability flow, so $\mathcal A(b)=0$. Equivalently, each Euler update reproduces exactly the increment of the affine trajectory. Hence, the straight path is integrated exactly by Euler on any time grid.

This exactness property raises a natural question: beyond the straight path, what constraints does exact finite-step Euler sampling impose on the drift?
For a uniform $N$-step Euler scheme with $h=1/N$, the Gaussian dynamics decouple coordinatewise as
\[
\widetilde X_{k+1,i}
=
\bigl(1+h a_{k/N,i}\bigr)\widetilde X_{k,i}.
\]
Hence, defining
\(
q_i
=
\prod_{k=0}^{N-1}
\bigl(1+h a_{k/N,i}\bigr),
\)
the terminal variance in direction $u_i$ is $c_i q_i^2$, and therefore, 
\begin{equation}\label{eq:w2}
W_2^2\!\left(\Law(X_1),\Law(\widetilde X_1)\right)
=
\sum_{i=1}^D
c_i\left(\sqrt{\rho_i}-|q_i|\right)^2.
\end{equation}
Therefore, exactness requires the product of its stepwise factors to reproduce the prescribed terminal variance, namely $\left|\prod_{k=0}^{N-1}(1+ha_{k/N})\right|=\sqrt\rho$. This yields a necessary lower bound on the drift magnitude along the discretization grid.

\begin{theorem}[Finite-step drift bound]\label{thm:stiffness}
Fix a direction with variance ratio $\rho$. If the uniform $N$-step Euler scheme produces the exact terminal law in this direction, then
\begin{equation}\label{eq:bound}
\max_{0\le k<N}|a_{k/N}|
\ge
N\big|\rho^{1/(2N)}-1\big|.
\end{equation}
For $N=1$, the lower bound is $|\sqrt{\rho}-1|$, while as $N\to +\infty$, it converges to $
\frac12|\log\rho|$.
\end{theorem}
The lower bound decreases with $N$ when $\rho>1$ and increases with $N$ when $0<\rho<1$. At $N=1$, it coincides with the drift magnitude of the Wasserstein geodesic, which is integrated exactly by Euler. As $N\to\infty$, it converges to $\frac12|\log\rho|$, the minimum of $\sup_t | a_t |$ among continuous paths with the prescribed endpoint variances. Thus, the bound connects the one-step exactness scale with the continuous-time minimal-drift scale.

The previous bound is only a necessary condition: attaining it does not, in general, guarantee exact integration under Euler. Under mild assumptions on the path, however, finite-step exactness becomes more restrictive and singles out the geodesic.

\begin{proposition}[Exactness of stable convex paths]\label{prop:convex} Fix one eigendirection. Suppose $\varphi_t$ is convex and the Euler updates are stable, i.e., $1+h a_{k/N}\ge 0$, $k=0,\ldots,N-1$.
If the Euler scheme produces the exact terminal law for some finite $N$, then $\varphi_t$ is affine. Consequently, the geodesic path~\eqref{eq:straight} is the unique convex path with stable Euler updates that can be integrated exactly at finite step size.
\end{proposition}

Together, Theorem~\ref{thm:stiffness} and Proposition~\ref{prop:convex} highlight the distinction between continuous-time regularity and finite-step accuracy. The logarithmic path minimizes the drift magnitude, whereas the straight path is distinguished by its compatibility with Euler discretization: its affine flow is exact on any grid and, within the convex class above, is the only path with this property.

\section{The factorization controls the conditional regression variance}\label{sec:factor}

We now turn to the second degree of freedom in the schedule. Once the variance path $(r_t)$ is fixed, \Cref{prop:separation} shows that the mixing angle $\theta_t$ leaves the marginal path and probability-flow drift unchanged,
and controls the conditional regression variance,
$
\operatorname{Var}(\dot I_{t,i}\mid I_{t,i})
=c_i r_{t,i}\dot\theta_{t,i}^2.
$
The factorization can therefore be optimized for training without altering the sampling dynamics. Two objectives then arise: minimizing $\int_{0}^1 \operatorname{Var}(\dot I_{t,i}\mid I_{t,i}) \mathrm{d}t$, or making $\operatorname{Var}(\dot I_{t,i}\mid I_{t,i})$ uniform in time. Both admit explicit solutions.

\begin{corollary}[Optimal angle design]\label{cor:angles}
In direction $u_i$, among angles with $\theta_0=0$ and $\theta_1=\pi/2$:
\begin{enumerate}
\item The expected conditional variance over uniformly-sampled time is minimized by
\begin{equation}\label{eq:averageangle}
\dot\theta_t
=\frac{\pi}{2}\frac{r_t^{-1}}{\int_0^1r_s^{-1}\mathrm{d}s},
\end{equation}
\item Among nondecreasing angles, the conditional variance is constant in time if and only if
\begin{equation}\label{eq:constantangle}
\dot\theta_t
=\frac{\pi}{2}\frac{r_t^{-1/2}}{\int_0^1r_s^{-1/2}ds}.
\end{equation}
\end{enumerate}
\end{corollary}

The proof is given in Appendix~\ref{app:angleproof}.
The two designs redistribute the angular motion differently along the same variance path: \eqref{eq:averageangle} satisfies $\dot\theta_t\propto r_t^{-1}$, so the angle changes faster where $r_t$ is small. In contrast, constant conditional variance \eqref{eq:constantangle} requires $\dot\theta_t\propto r_t^{-1/2}$ to compensate for the factor $r_t$ in $\operatorname{Var}(\dot I_{t,i}\mid I_{t,i})=c_i r_{t,i}\dot\theta_t^2$. For the geodesic and the logarithmic path, both angle choices admit closed-form expressions; see Appendix~\ref{app:closedformangles}.

Both choices differ only in how $X_0$ and $X_1$ are combined along the interpolation. Through~\eqref{eq:polar}, each choice defines a complete stochastic interpolant schedule. Lower or more uniform regression variance does not, by itself, guarantee faster optimization in practice: gradient variance and learned-model error also depend on the model class, parametrization, and optimizer. Note that a non-uniform distribution of training times simply reweights the corresponding time-averaged variational problem.

\section{Discussion}
Our Gaussian analysis separates two design choices within a schedule for stochastic interpolants: the variance path determines sampling, whereas its factorization controls regression variance. This suggests choosing the path for numerical sampling and then its factorization for training, without altering the probability flow. Our exact results rely on centered, commuting Gaussians and direction-dependent schedules; beyond this setting, the decomposition should be viewed as a covariance-informed design principle rather than an exact characterization.

\section*{Acknowledgments}
We thank Giulio Biroli for interesting discussions.

\bibliographystyle{plainnat}
\bibliography{refs}

\clearpage
\appendix
\section{Proofs and supporting calculations}

\subsection{Proof of proposition \ref{prop:separation}}\label{app:separation}

The coordinates of $(I_t,\dot I_t)$ are independent across $i$ in the common basis, so conditioning the $i$th coordinate on $I_t$ is equivalent to conditioning it on $I_{t,i}$. Moreover, $(\dot I_{t,i},I_{t,i})$ is jointly centered Gaussian, therefore Gaussian conditioning gives
\[
b_t(x)_i
= \E[\dot I_{t,i}\mid I_{t,i}=x_i]
=\frac{\operatorname{Cov}(\dot I_{t,i},I_{t,i})} {\operatorname{Var}(I_{t,i})}\,x_i
= \frac{1}{2}\frac{\dot r_{t,i}}{r_{t,i}}\,x_i,
\]
which proves \eqref{eq:drift}.

Let $z_t=(\alpha_{t,i},\sqrt{\rho_i}\beta_{t,i})$. Gaussian conditioning also gives
\[
\operatorname{Var}(\dot I_{t,i}\mid I_{t,i})
= \operatorname{Var}(\dot I_{t,i})
-\frac{\operatorname{Cov}(\dot I_{t,i},I_{t,i})^2}
     {\operatorname{Var}(I_{t,i})}
=c_i\left(|\dot z_t|^2-\frac{\dot r_{t,i}^2}{4r_{t,i}}\right).
\]
The polar form yields $|\dot z_t|^2=\dot r_{t,i}^2/(4r_{t,i})+r_{t,i}\dot\theta_{t,i}^2$, proving \eqref{eq:variance}.

\subsection{The two path optima}
Fix one direction and use the notation of
Section~\ref{sec:path}: $r_t=r_{t,i}$, $\rho=\rho_i$, and
$\varphi_t=\sqrt{r_t}$. Proposition~\ref{prop:separation} gives $a_t=\dot\varphi_t/\varphi_t$ and $\operatorname{Var}(I_{t,i})=c_i\varphi_t^2$. Hence
\[
\int_0^1\E|b_t(I_t)_i|^2\mathrm{d}t
=c_i\int_0^1\dot\varphi_t^2 \mathrm{d}t.
\]
Since $\varphi_0=1$ and $\varphi_1=\sqrt{\rho}$,
Cauchy--Schwarz gives
\[
\int_0^1\dot\varphi_t^2\mathrm{d}t
\ge \left(\int_0^1 \dot \varphi_t\,\mathrm{d} t \right)^2
=(\sqrt{\rho}-1)^2,
\]
with equality if and only if $\dot\varphi_t$ is constant. This gives
\eqref{eq:straight}. 
Likewise,
\[
\int_0^1|a_t|^2 \mathrm{d}t
=\int_0^1 \left(\partial_t\log\varphi_t\right)^2 \mathrm{d}t
\ge 
\left( \int_0^1\partial_t\log\varphi_t\,\mathrm{d} t \right)^2
=\frac14(\log\rho)^2,
\]
with equality if and only if $\log\varphi_t$ is affine, which gives \eqref{eq:logpath}.

\subsection{Euler error bound}\label{app:eulerbound}
Let $h=1/N$, $t_k=kh$, and let $\widetilde X_{t_k}$ denote the
Euler approximation coupled with the exact solution through the same
initial condition, namely: 
\[
\widetilde X_{t_0}=X_0,
\qquad
\widetilde X_{t_{k+1}} = \widetilde X_{t_k} + h b_{t_k}(\widetilde X_{t_k}).
\]
Define the material derivative by $ D_t b_t = \partial_t b_t+\nabla_x b_t\,b_t. $
Since $\frac{\mathrm{d}}{\mathrm{d}s}b_s(X_s)=D_sb_s(X_s)$, the local defect is
\[
\delta_k:=X_{t_{k+1}}-X_{t_k}-h b_{t_k}(X_{t_k})
=\int_{t_k}^{t_{k+1}}(t_{k+1}-s)D_sb_s(X_s)\,ds.
\]
Combining the first and last equations yields: 
\[
X_{t_{k+1}}-\widetilde X_{t_{k+1}}
= X_{t_k}-\widetilde X_{t_k} + h\bigl( b_{t_k}(X_{t_k})-b_{t_k}(\widetilde X_{t_k}) \bigr) + \delta_k.
\]

Set $e_k=(\E\|X_{t_k}-\widetilde X_{t_k}\|^2)^{1/2}.$ The Lipschitz assumption, Minkowski's inequality, and
Cauchy--Schwarz give
\[
e_{k+1}\le(1+hL)e_k+\|\delta_k\|_{L^2},
\qquad 
\|\delta_k\|_{L^2}^2
\le\frac{h^3}{3}\int_{t_k}^{t_{k+1}}\E\|D_sb_s(X_s)\|^2ds.
\]

Since $e_0=0$, iterating the recursion and using $(1+hL)^N\le e^L$ yields
\[
\begin{aligned}
e_N \le e^L\sum_{k=0}^{N-1}\lVert\delta_k\rVert_{L^2}
&\le e^L \left( N\sum_{k=0}^{N-1}\lVert\delta_k\rVert_{L^2}^2 \right)^{1/2}\\
& \le e^L \left(\frac1h \frac{h^3}{3}\sum_{k=0}^{N-1} \int_{t_k}^{t_{k+1}} \mathbb E\| D_s b_s(X_s)\|^2 \mathrm{d}s \right)^{1/2}\\
&\le \frac{e^Lh}{\sqrt{3}}\, \mathcal{A}(b)^{1/2},
\end{aligned}
\]
Finally, the chosen coupling gives $ W_2\bigl(\Law(X_1),\Law(\widetilde X_1)\bigr) \le e_N, $ which proves \eqref{eq:eulererror}.

\subsection{Proof of Corollary~\ref{cor:angles}}\label{app:angleproof}

By \eqref{eq:variance}, minimizing the time-averaged conditional variance amounts to minimizing $\int_0^1 r_t\dot\theta_t^2\,\mathrm{d}t$.
Cauchy--Schwarz gives
\[
\left(\frac\pi2\right)^2
=\left(\int_0^1\dot\theta_t\mathrm{d}t\right)^2
\le\left(\int_0^1r_t\dot\theta_t^2\mathrm{d}t\right)
\left(\int_0^1r_t^{-1}\mathrm{d}t\right),
\]
with equality exactly in \eqref{eq:averageangle}. Equation~\eqref{eq:constantangle} follows by requiring  $r_t\dot\theta_t^2$ to be constant and applying the endpoint constraint.

\subsection{Closed-form angles for the two optimal paths}
\label{app:closedformangles}
Fix one direction with $\rho\ne1$. Since $\theta_0=0$,
Equations~\eqref{eq:averageangle} and~\eqref{eq:constantangle} give
\[
\theta_t^{\mathrm{avg}}
=\frac{\pi}{2}
\frac{\int_0^t r_s^{-1}\,\mathrm{d}s}
     {\int_0^1 r_s^{-1}\,\mathrm{d}s},
\qquad
\theta_t^{\mathrm{const}}
=\frac{\pi}{2}
\frac{\int_0^t r_s^{-1/2}\,\mathrm{d}s}
     {\int_0^1 r_s^{-1/2}\,\mathrm{d}s}.
\]

For the straight Gaussian Wasserstein path,
$r_t=(1+(\sqrt\rho-1)t)^2$. Direct integration gives
\[
\int_0^t r_s^{-1}\,\mathrm{d}s 
=\frac{t}{1+(\sqrt\rho-1)t},
\qquad
\int_0^t r_s^{-1/2}\,\mathrm{d}s 
=\frac{\log(1+(\sqrt\rho-1)t)}{\sqrt\rho-1}.
\]
Normalizing these expressions at $t=1$ yields
\[
\theta_t^{\mathrm{avg}}
=\frac{\pi}{2}\frac{\sqrt\rho\,t}{1+ (\sqrt\rho-1)t},
\qquad
\theta_t^{\mathrm{const}}
=\frac{\pi}{2} \frac{\log(1+(\sqrt\rho-1)t)}{\log(\sqrt\rho)}.
\]

For the logarithmic path, $r_t=\rho^t$. In this case,
\[
\int_0^t r_s^{-1}\,\mathrm{d}s
=\frac{1-\rho^{-t}}{\log\rho},
\qquad
\int_0^t r_s^{-1/2}\,\mathrm{d}s
=\frac{2(1-\rho^{-t/2})}{\log\rho}.
\]
Normalizing again at $t=1$ gives
\[
\theta_t^{\mathrm{avg}}
=\frac{\pi}{2}\frac{1-\rho^{-t}}{1-\rho^{-1}},
\qquad
\theta_t^{\mathrm{const}}
=\frac{\pi}{2}\frac{1-\rho^{-t/2}}{1-\rho^{-1/2}}.
\]

\subsection{Proof of theorem \ref{thm:stiffness}} \label{app:stiffness}

Let $M=\max_k|a_{k/N}|$. Exactness in law requires
\[
\left|\prod_{k=0}^{N-1}\left(1+\frac{a_{k/N}}{N}\right)\right|
=\sqrt{\rho}.
\]
If $\rho\ge 1$, the triangle inequality gives
\[
\sqrt{\rho}
\le
\prod_{k=0}^{N-1}\left(1+\frac{|a_{k/N}|}{N}\right)
\le
\left(1+\frac{M}{N}\right)^N,
\]
and hence $M\ge N(\rho^{1/(2N)}-1)$.

Now suppose $0<\rho<1$. If $M\ge N$, then \eqref{eq:bound} holds immediately, since $N(1-\rho^{1/(2N)})<N$. If $M<N$, every Euler factor is positive and satisfies
\[
1+\frac{a_{k/N}}{N}\ge 1-\frac{M}{N}.
\]
Therefore,
\[
\sqrt{\rho}
=
\prod_{k=0}^{N-1}\left(1+\frac{a_{k/N}}{N}\right)
\ge
\left(1-\frac{M}{N}\right)^N,
\]
which gives $M\ge N(1-\rho^{1/(2N)})$. Combining the two cases proves \eqref{eq:bound}. The values at $N=1$, the limit, and the stated monotonicity follow from elementary properties of the exponential function.

\subsection{Proof of Proposition~\ref{prop:convex}}
Convexity gives
\[
\varphi(t_k)+h\dot\varphi(t_k)\le\varphi(t_{k+1}),
\quad\text{or}\quad
\varphi(t_k)(1+h a_{t_k})\le\varphi(t_{k+1}).
\]
Let $y_0=\varphi(0)=1$ and $y_{k+1}=y_k(1+h a_{t_k})$. Stability preserves the inequality, so induction gives $0\le y_k\le\varphi(t_k)$. If the final law is exact, then $y_N=\varphi(1)>0$ and every preceding inequality must be an equality. A differentiable convex function whose tangent at each $t_k$ meets its graph at $t_{k+1}$ is affine on every grid interval, hence on $[0,1]$.

As a consequence, any stable finite-step exact path that is not
straight must be nonconvex and hence satisfy $\ddot\varphi_t<0$, or equivalently $\dot a_t+a_t^2<0$, on a set of positive measure.

\subsection{Optimal schedulers for a mixed variational problem}
\label{app:combination_obj}
In this appendix, we consider the following mixed variational problem:
\begin{equation}
\label{eq:variational_form}
\mathcal{B}(b):=\min\limits_{r_t}
\int_0^1\mathbb{E}\big[\|b_t(I_t)\|_2^2\big]\,\mathrm{d}t
+\lambda\int_0^1\mathbb{E}\big[\|\nabla_x b_t(I_t)\|_F^2\big]\,\mathrm{d}t
=
\sum\limits_{i=1}^D
\int_0^1
\frac{1}{4}
\left(
\frac{c_i\dot r_{t,i}^2}{r_{t,i}}
+\lambda\frac{\dot r_{t,i}^2}{r_{t,i}^2}
\right)\mathrm{d}t, \quad \lambda>0
\end{equation}
This objective complements the Optimal Transport and Lipschitz objectives considered in \citep{cvx2025}, which correspond, respectively, to the limiting cases $\lambda=0$ and $\lambda=+\infty$.

Since the eigen-directions are independent of one another, the problem can be treated separately along each direction. The solution to \eqref{eq:variational_form} is characterized by the Euler--Lagrange equations, which can be written:
\begin{equation}
\label{eq:fixed_point_equation}
F_\lambda(r_{t,i})
:=2\big(\sqrt{c_ir_{t,i}+\lambda}-\sqrt{c_i+\lambda}\big)+\sqrt{\lambda}\,\log\!\left(
\frac{(\sqrt{c_ir_{t,i}+\lambda}-\sqrt{\lambda})
(\sqrt{c_i+\lambda}+\sqrt{\lambda})}{(\sqrt{c_ir_{t,i}+\lambda}+\sqrt{\lambda})
(\sqrt{c_i+\lambda}-\sqrt{\lambda})}\right)=\gamma_i t,
\end{equation}
where $\gamma_i$ is an integration constant that depends on the direction under consideration. Moreover, $\operatorname{sign}(\gamma_i)=\operatorname{sign}(\log(\rho_i)),$ and by construction $\gamma_i=F_\lambda(\rho_i).$
The fixed-point equation \eqref{eq:fixed_point_equation} admits a unique solution for each $t$, although not available in closed-form. Indeed, $F_\lambda$ is continuous on $\mathbb{R}_+^*$ with derivative
$F_\lambda'(r_{t,i})=\frac{\sqrt{\lambda+c_ir_{t,i}}}{r_{t,i}} >0,$
so that $F_\lambda$ is strictly increasing. Furthermore,
$F_\lambda(0^+)=-\infty, \, F_\lambda(1)=0$.
Hence, for $t\in[0,1]$, there exists a unique $r_{t,i}$ satisfying $F_\lambda(r_{t,i})=\gamma_i t.$ By differentiating \eqref{eq:fixed_point_equation} with respect to $t$, one shows that the velocity field linear coefficient is 
$$
a_{i,t}:=\frac{\dot r_{t,i}}{2r_{t,i}}=\frac{\gamma_i}2\frac1{\sqrt{\lambda+c_ir_{t,i}}}.
$$

Newton's method is an appealing approach for solving \eqref{eq:fixed_point_equation} due to its quadratic convergence rate \citep{dennis1996numerical}.
However, the method requires an initial guess sufficiently close to the solution to achieve this convergence rate. To obtain a suitable initialization, we therefore consider an approximate version of the problem.

\begin{figure}[ht]
    \centering
    \includegraphics[width=.9\linewidth]{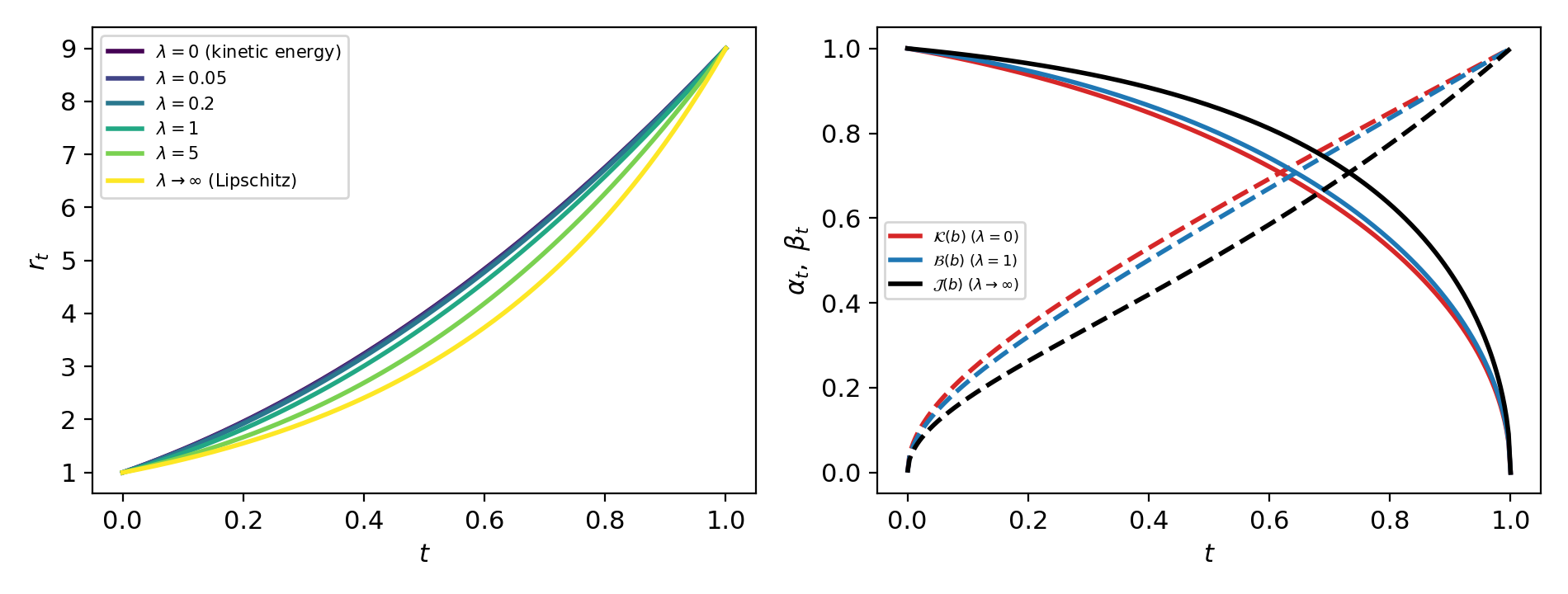}
    \caption{\textit{Left}: Optimal eigenvalue $(r_{t})_{t\in[0, 1]}$ for different regularization strength. \textit{Right}: Optimal schedulers $(\alpha_t)_{t\in[0, 1]}$ (continuous) and $(\beta_t)_{t\in[0, 1]}$ (dashed) for the three objectives $\mathcal{K}, \mathcal{B}$ and $\mathcal{J}$.}
    \label{fig:optimal_eigen_path}
\end{figure}

Using the approximation
$\log(1+x)=\frac{x}{\sqrt{1+x}}+\mathcal{O}(x^3)$ when $x$ is close to $0$,
we approximate the logarithmic term in \eqref{eq:fixed_point_equation} by choosing
$x=\frac{-2\sqrt{\lambda}}{\sqrt{\lambda+c_ir_{t,i}}+\sqrt{\lambda}},$
which gives $1+x=\frac{\sqrt{\lambda+c_ir_{t,i}}-\sqrt{\lambda}}
{\sqrt{\lambda+c_ir_{t,i}}+\sqrt{\lambda}}.$
Substituting this into \eqref{eq:fixed_point_equation} and collecting the constant terms on the right-hand side yields
\[
\sqrt{c_ir_{t,i}+\lambda}-\frac{\lambda}{\sqrt{c_ir_{t,i}}}=\frac12\gamma_i t+\kappa_i :=s_t.
\]
The constant $\kappa_i$ is determined by the endpoint condition $t=0$, $\kappa_i=\sqrt{c_i+\lambda}-\frac{\lambda}{\sqrt{c_i}}.$ After moving the right term to the rhs, and squaring both sides, we obtain the following depressed quartic equation:
$$
y_t^4 + (\lambda-s_t^2)y_t^2 - 2s_t\lambda y_t - \lambda^2=0,
$$
in the variable $y_t=\sqrt{c_ir_{t,i}}$.
It can be solved in radicals using Ferrari's method \citep{girstmair2026solving}. Among its roots, we select the smallest positive real root as the initial guess and then run Newton's algorithm.  Once $r_{t,i}$ has been determined, an additional degree of freedom, $\theta_{i,t}$, remains undetermined. In \citep{cvx2025}, the angle trajectory is implicitly fixed when imposing additionally $\alpha_{i,t}^2=1-\beta_{i,t}^2$ for all $t\in[0,1]$. Figure \ref{fig:optimal_eigen_path} illustrates the resulting trajectories of $r_t$, $\alpha_t$, and $\beta_t$ for $\rho=9$. 

\end{document}